\documentclass[conference, 9pt]{IEEEtran}

\usepackage[normalem]{ulem}
\usepackage{amsmath}
\usepackage{amsfonts}
\usepackage{hyperref}
\usepackage{graphicx}
\usepackage{cleveref}
\usepackage{amsthm}
\usepackage{algorithm, algpseudocode,subcaption}
\usepackage{xcolor}
\usepackage{tikz}
\usepackage{caption}
\usepackage{listings}
\usepackage{array}
\usepackage{booktabs}
\usepackage{diagbox}
\usepackage{float}
\usepackage{geometry}
\newtheorem{theorem}{Theorem}
\newtheorem{lemma}{Lemma}

\newtheorem{corollary}{Corollary}
\newtheorem{definition}{Definition}

\title{Robust Anomaly Detection via Tensor Pseudoskeleton Decomposition}
\author{Bowen Su }
\definecolor{officegreen}{rgb}{0.0, 0.5, 0.0}
\begin{document}

\maketitle

\begin{abstract}
Anomaly detection plays a critical role in modern data-driven applications, from identifying fraudulent transactions and safeguarding network infrastructure to monitoring sensor systems for irregular patterns. Traditional approaches—such as distance-, density-, or cluster-based methods, face significant challenges when applied to high-dimensional tensor data, where complex interdependencies across dimensions amplify noise and computational complexity. To address these limitations, this paper leverages Tensor  pseudoskeleton decomposition within a tensor-robust principal component analysis framework to extract low-Tucker-rank structure while isolating sparse anomalies, ensuring robustness to anomaly detection. We establish theoretical analysis of convergence, and estimation error, demonstrating the stability and accuracy of the proposed approach. Numerical experiments on real-world spatiotemporal data from New York City taxi trip records validate the superiority of the proposed method in detecting anomalous urban events compared to existing benchmark methods. The results underscore the potential of Tensor  pseudoskeleton decomposition to enhance anomaly detection for large-scale, high-dimensional data.
\end{abstract}
\section{Introduction}
Anomaly detection is a crucial task in data analysis, with applications spanning various domains such as fraud detection~\cite{motie2024financial}, cybersecurity~\cite{wurzenberger2024analysis}, healthcare monitoring~\cite{kadir2024anomaly}, and sensor network analysis~\cite{tarish2025anomaly}. Anomalies, or outliers, represent data points or patterns that deviate significantly from the expected behavior, often signaling critical events or errors that require immediate attention. Detecting these anomalies, especially within high-dimensional and complex datasets, is challenging due to the sheer volume of data and the underlying noise that can mask unusual patterns.

Traditional anomaly detection techniques, including distance-based~\cite{Angiulli2002}, density-based~\cite{Breunig2000}, and clustering-based methods~\cite{Jiang2003,Hautamaki2004}, have shown some success in identifying anomalies in lower-dimensional datasets.
 However, these approaches often struggle when extended to high-dimensional tensor data, where intricate dependencies exist across multiple dimensions. Tensor data structures are common in fields such as video surveillance, biomedical imaging, and environmental monitoring, where data is naturally organized in multi-way arrays. The increased dimensionality not only complicates the detection of anomalies but also amplifies the computational costs, making scalability a critical concern.

In recent years, tensor decomposition methods have emerged as powerful tools for managing high-dimensional data. By transforming complex data into a lower-dimensional, interpretable form, tensor decompositions facilitate efficient storage, processing, and analysis. Among these methods, Tucker decomposition, a form of higher-order singular value decomposition, is particularly effective at capturing the core structure of tensor data. However, while Tucker decomposition enables significant dimensionality reduction, it remains sensitive to outliers, which can distort the decomposition and lead to unreliable results in anomaly detection.

To address these limitations,  Tensor  pseudoskeleton decomposition offers an alternative approach by selecting representative parts of the data, thereby preserving essential features while reducing redundancy.  Tucker pseudoskeleton decomposition provides a structured decomposition that is both computationally efficient and robust~\cite{hamm2023generalized,cai2021mode}. 

In this paper, we focus on anomaly detection within the tensor robust principal component analysis framework by leveraging a Tucker pseudoskeleton decomposition specifically tailored for high-dimensional datasets~\cite{hamm2023generalized,cai2021mode}. By incorporating sparsity and regularization constraints, our method reduces sensitivity to anomalies, enabling more accurate and resilient detection of unusual patterns. The Tucker  pseudoskeleton decomposition framework combines the strengths of Tucker decomposition’s structural insight with pseudoskeleton’s selective feature extraction while enhancing robustness against outliers~\cite{hamm2023generalized}.

\subsection{Notations and definitions}
In this section, we introduce notation and review foundational properties of Tucker-based tensor decomposition, which will be essential throughout the chapter. Tucker decomposition serves as a powerful tool for capturing the core structure of high-dimensional data, providing both a compact representation and interpretability of multi-dimensional relationships within the data.

To distinguish between different mathematical entities, we adopt the following conventions: calligraphic capital letters (e.g., $\mathcal{T}$) represent tensors, regular uppercase letters (e.g., ${X}$) denote matrices, regular lowercase letters (e.g., ${x}$) indicate vectors or scalars. For submatrices, $[X]_{I,:}$ and $[X]_{:,J}$ refer to the rows and columns of matrix ${X}$ indexed by sets $I$ and $J$, respectively. For tensors, $[\mathcal{T}]_{I_1, \dots, I_n}$ represents a subtensor of $\mathcal{T}$ with index sets $I_k$ along each mode $k$. A specific element in a tensor is accessed by the index notation $[\mathcal{T}]_{i_1, \dots, i_n}$. 

The tensor norm used in this chapter is the Frobenius norm~\cite{kolda2009tensor}, defined for a tensor \(\mathcal{T}\) as:
\begin{equation*}
    \|\mathcal{T}\|_\mathrm{F} = \sqrt{\sum_{i_1, \dots, i_n} [\mathcal{T}]_{i_1, \dots, i_n}^2}.
\end{equation*}
This norm represents the square root of the sum of the squared entries of \(\mathcal{T}\), extending the Frobenius norm from matrices to higher-order tensors.
For matrices, the Moore-Penrose Pseudoinverse is denoted by ${X}^\dagger$. The notation $[d] := \{1, \dots, d\}$ represents the set of natural numbers up to $d$.

\begin{definition}[\textbf{Tensor Matricization/Unfolding}~\cite{kolda2009tensor}]
    An $n$-mode tensor $\mathcal{T}$ can be reshaped into a matrix by unfolding it along each of its $n$ modes. The mode-$k$ unfolding of a tensor $\mathcal{T} \in \mathbb{R}^{d_1 \times \dots \times d_n}$, denoted $\mathcal{T}_{(k)}$, is a matrix of size $\mathbb{R}^{d_k \times \prod_{j \neq k} d_j}$, obtained by arranging all vectors of $\mathcal{T}$ with indices fixed in all modes except the $k$-th. This transformation, $\mathcal{T} \mapsto \mathcal{T}_{(k)}$, is referred to as the mode-$k$ unfolding operator.
\end{definition}

\begin{definition}[\textbf{Mode-$k$ Product}~\cite{kolda2009tensor}]
    Let $\mathcal{T} \in \mathbb{R}^{d_1 \times \dots \times d_n}$ and ${A} \in \mathbb{R}^{J \times d_k}$. The mode-$k$ product of $\mathcal{T}$ with ${A}$, denoted by $\mathcal{Y} = \mathcal{T} \times_k {A}$, is defined element-wise as:
    \begin{equation*}
        [\mathcal{Y}]_{i_1, \dots, i_{k-1}, j, i_{k+1}, \dots, i_n} = \sum_{s=1}^{d_k} [\mathcal{T}]_{i_1, \dots, i_{k-1}, s, i_{k+1}, \dots, i_n} [{A}]_{j, s}.
    \end{equation*}
    Alternatively, this operation can be represented in matrix form as $\mathcal{Y}_{(k)} = {A} \mathcal{T}_{(k)}$. For a sequence of tensor-matrix products across different modes, we use the notation $\mathcal{T} \times_{i=t}^{s} {A}_i$ to indicate the product $\mathcal{T} \times_{t} {A}_{t} \times_{t+1} \dots \times_{s} {A}_{s}$. This operation is referred to as the `tensor-matrix product' throughout the paper.
\end{definition}
\begin{definition}[\textbf{Tucker Rank and Tucker Decomposition}~\cite{kolda2009tensor}]
    The Tucker decomposition of a tensor $\mathcal{T}$ approximates it by expressing it as a product of a core tensor $\mathcal{C}$ and factor matrices ${A}_k$ along each mode:
    \begin{equation*}\label{eqn:Tucker_Decomposition}
        \mathcal{T} \approx \mathcal{C} \times_{i=1}^n {A}_i.
    \end{equation*}
    If the approximation in \eqref{eqn:Tucker_Decomposition} becomes an equality and the core tensor $\mathcal{C} \in \mathbb{R}^{r_1 \times \dots \times r_n}$, this is termed an exact Tucker decomposition of $\mathcal{T}$. The ranks $(r_1, \dots, r_n)$ are known as the Tucker ranks of the tensor $\mathcal{T}$.
\end{definition}

In the realm of matrix algebra, the pseudoskeleton decomposition technique is a good alternative to SVD~\cite{Goreinov}. Specifically, this method entails selecting specific columns ${C}$ and rows ${R}$ from a matrix ${X} \in \mathbb{R}^{d_1 \times d_2}$, and constructing a core matrix ${U} = {X}(I, J)$. The matrix ${X}$ is then reconstructed through the product ${C} {U}^\dagger {R}$, under the condition that $\operatorname{rank}({U}) = \operatorname{rank}({X})$. Expanding from matrices to tensors, the initial adaptations of pseudoskeleton decompositions applied a single-mode unfolding to 3-mode tensors~\cite{mahoney2008tensor}. To my best knowledge, the following are recent works on tensor pseudoskeleton decompositions or Tensor CUR Decompostions~\cite{hamm2023generalized,cai2021mode,cai2023robust,ahmadi2022cross,caiafa2010generalizing}. Furthermore, H. Cai, K. Hamm, and etc have presented rigorous theoretical results on the exact tensor pseudoskeleton decomposition \cite{cai2021mode,cai2023robust,hamm2023generalized}. For completeness, we present their work below.
\begin{definition}[Tensor pseudoskeleton decompositions or Tensor CUR Decompostions~\cite{hamm2023generalized,cai2021mode,cai2023robust}]
 Consider a tensor \(\mathcal{A} \in \mathbb{R}^{d_1 \times \cdots \times d_n}\) with Tucker ranks \((r_1, \dots, r_n)\). Suppose that for each mode \(i\), there exists a subset \(I_i \subseteq [d_i]\) such that 
 \[
\mathcal{A} = \mathcal{R} \times_{i=1}^{n} \left( C_i U_i^\dagger \right),
\]
where \(\mathcal{R} = [\mathcal{A}]_{I_1, \dots, I_n}, \quad C_i = [\mathcal{A}_{(i)}]_{:, J_i}, \quad \text{and} \quad U_i = [\mathcal{C}_{(i)}]_{I_i, :}, \quad J_i = \bigotimes_{j \neq i} I_j.
\)
\end{definition}
\begin{theorem}[{~\cite{hamm2023generalized,cai2021mode,cai2023robust}}]\label{thm:  Tucker Decomposition}
    For a tensor $\mathcal{A} \in \mathbb{R}^{d_1 \times \cdots \times d_n}$ with Tucker ranks $(r_1, \dots, r_n)$, consider subsets $I_i \subseteq [d_i]$ and let $J_i = \bigotimes_{j \neq i} I_j$ for each mode $i$. Define $\mathcal{R} = [\mathcal{A}]_{I_1, \dots, I_n}$, ${C}_i = [\mathcal{A}_{(i)}]_{:, J_i}$, and ${U}_i = [\mathcal{C}_{(i)}]_{I_i, :}$. The following conditions are equivalent:
    \begin{enumerate}
        \item $\mathcal{A} = \mathcal{R} \times_{i=1}^{n} ({C}_{i} {U}_i^\dagger)$,
        \item $\operatorname{rank}({U}_i) = r_i$ for all $i$,
        \item $\operatorname{rank}({C}_i) = r_i$ for all $i$, and $\mathcal{R}$ has Tucker rank $(r_1, \dots, r_n)$.
    \end{enumerate}
\end{theorem}
 For those interested in further details, It is recommended to read works~\cite{hamm2023generalized,cai2021mode,ahmadi2022cross,caiafa2010generalizing,cai2023robust,che2022perturbations,saibaba2016hoid}.

\section{Methodology}
We employ Tensor Robust Principal Component Analysis, an extension of classical Robust PCA that can operate directly on multi-dimensional (tensor) data. Unlike conventional low-rank models that assume the entire dataset is low-rank, TRPCA decomposes a given tensor into two distinct components: a low-rank component representing regular patterns and a sparse component isolating anomalies. This decomposition effectively isolates outliers in spatial-temporal data while retaining core structural patterns, providing a more flexible and robust approach to anomaly detection. By handling high-dimensional tensor data, TRPCA is particularly well-suited for scenarios where data is naturally structured as a multi-way array, allowing for the detection of unusual patterns that vary across both space and time.

In this framework, we represent the spatial-temporal data as a tensor $\mathcal{T} \in \mathbb{R}^{d_1 \times d_2 \times \cdots \times d_n}$, where each dimension $d_i$ corresponds to a specific mode of the data. For example, $d_1$ might represent spatial coordinates, $d_2$ temporal intervals, and additional dimensions might capture contextual features or sensor types. The objective is to decompose $\mathcal{T}$ into two components: a low-rank tensor $\mathcal{L}^\star$ that captures the dominant spatial-temporal structure, and a sparse tensor $\mathcal{S}^\star$ representing anomalies or outliers. The decomposition is expressed as:
\begin{equation*}
    \mathcal{T} = \mathcal{L}^\star + \mathcal{S}^\star,
\end{equation*}
where $\mathcal{L}^\star \in \mathbb{R}^{d_1 \times \cdots \times d_n}$ encapsulates the smooth, regular patterns in the data, while $\mathcal{S}^\star \in \mathbb{R}^{d_1 \times \cdots \times d_n}$ captures deviations from these patterns, isolating events that significantly differ from expected behavior. This separation allows for robust anomaly detection, as $\mathcal{S}^\star$ can pinpoint localized irregularities without interference from the regular structure. Mathematically, we formulate the anomaly detection problem as an optimization problem that seeks to minimize the reconstruction error between $\mathcal{T}$ and the sum of $\mathcal{L}$ and $\mathcal{S}$. This is achieved through the following objective:
\begin{equation*}\label{eq:trpca_formulation}
    \begin{split}
        \min_{\mathcal{R}, {C}_{i}, {U}_i, \mathcal{S}} & \quad \|\mathcal{T} - \mathcal{L} - \mathcal{S}\|_\mathrm{F} \\
        \text{subject to} & \quad \mathcal{L}= { \mathcal{R} \times_{i=1}^{n} ({C}_{i} {U}_i^\dagger)}\\
        &\quad\|\mathcal{S}\|_{\infty} \leq \alpha.
    \end{split}
\end{equation*}

\subsection{TRPCA via Tensor 
Pseudoskeleton Decomposition}

\begin{algorithm}[H]
    \caption{TRPCA via Tensor 
Pseudoskeleton Decomposition}
    \label{alg:TCPD}
    \begin{algorithmic}[1]
        \State \textbf{Input: } $\mathcal{T}  \in \mathbb{R}^{d_1 \times \cdots \times d_n}$: observed tensor; 
        $(r_1, \cdots, r_n)$: estimated Tucker rank; 
        $\varepsilon$: targeted precision; 
        $\zeta^{(0)}, \gamma$: thresholding parameters; $\{|I_i|\}_{i=1}^n,\{|J_i|\}_{i=1}^n$: cardinalities for sample indices.
        \State  Uniformly sample the indices $\{I_i\}_{i=1}^n, \{J_i\}_{i=1}^n$ 
        \State \textbf{Initialization:} $\mathcal{L}^{(0)} = 0, \mathcal{S}^{(0)} = 0, k = 0$
        \While {$e^{(k)} > \varepsilon$}
            \State \textcolor{officegreen}{// Step (I): Updating $\mathcal{S}$}
            \State $\zeta^{(k+1)} = \gamma \cdot \zeta^{(k)}$ 
            \State $\mathcal{S}^{(k+1)} = \mathrm{HT}_{\zeta^{(k+1)}}(\mathcal{T} - \mathcal{L}^{(k)})$  
            \State \textcolor{officegreen}{// Step (II): Updating $\mathcal{L}$}
            \State $\mathcal{L}^{(k+1)} = [\mathcal{T} - \mathcal{S}^{(k+1)}]_{I_1, \cdots, I_n}$
            \For{$i = 1, \cdots, n$}
                \State $C_i^{(k+1)} = [(\mathcal{T} - \mathcal{S}^{(k+1)})_{(i)}]_{:, J_i}$ 
                \State $[Q,R] = \operatorname{qr}\left([C_i^{(k+1)}]_{I_i, :}^{\top}\right)$

             \State$\mathcal{L}^{(k+1)} = \mathcal{L}^{(k+1)} \times C_i^{(k+1)}[Q]_{:,:r}[R]_{:r,:}^{\dagger}$    
            \EndFor
            
            \State $k = k + 1$ 
        \EndWhile
        \State \textbf{Output: } $\mathcal{L}^{(k+1)}, \mathcal{S}^{(k+1)}$.
    \end{algorithmic}
\end{algorithm}

\subsubsection{Step~(I): Update Sparse Component $\mathcal{S}$} 
\label{sec:updateS}
In this step, we update the sparse component \(\mathcal{S}\) — which captures data outliers — using the technique described in \cite{cai2023robust,netrapalli2014non,cai2019accelerated}. Specifically, we apply an iterative decaying threshold within the hard thresholding operator \(\mathrm{HT}_\zeta\) paired with \(\gamma\), as described in \cite{cai2023robust,CaiR2024}.
The hard thresholding operator $\mathrm{HT}_\zeta$ is defined as follows:

\begin{equation*}
    [\mathrm{HT}_{\zeta}(\mathcal{T})]_{i_1,\cdots,i_n} =
    \begin{cases}
        [\mathcal{T}]_{i_1,\cdots,i_n}, & \quad |[\mathcal{T}]_{i_1,\cdots,i_n}| > \zeta; \\
        0,  & \quad \text{otherwise.}
    \end{cases}
\end{equation*}

This operator $\mathrm{HT}_\zeta$ effectively filters out entries with magnitudes less than or equal to $\zeta$, treating them as negligible. By applying this to the tensor $\mathcal{T}$, only values deemed significant (i.e., values exceeding the threshold) remain in the updated sparse component $\mathcal{S}$, thereby enhancing the sparsity of $\mathcal{S}$.
\subsubsection{Step~(II): Update Low-Tucker-rank Component $\mathcal{L}$}
\label{sec:updateL}
In this step, we aim to update the low-Tucker-rank component \(\mathcal{L}\), which models the structured, low-rank part of the data tensor via tensor pseudoskeleton decomposition. The update process is divided into two key stages: subspace identification and projective reconstruction.
To approximate the low-rank structure along each mode, we begin by extracting the mode-\(i\) fibers from the residual tensor \(\mathcal{T} - \mathcal{S}^{(k)}\), which represents the current estimate of the sparse component subtracted from the observed data tensor. The fibers are assembled into the matrix representation:
\[
C_i^{(k)} \in \mathbb{R}^{d_i \times |J_i|},
\]
where each column of \(C_i^{(k)}\) corresponds to a mode-\(i\) fiber indexed by a subset of indices \(J_i\). We select a subset of mode-\(i\) fibers indexed by \(I_i \subseteq \{1, \dots, d_i\}\) and perform an economy-size QR decomposition on the transposed submatrix formed by these selected fibers:
\begin{equation*}
    \left[C_i^{(k)}\right]_{I_i,:}^\top = Q R,
\end{equation*}
where \(Q \in \mathbb{R}^{|J_i| \times r_i}\) is a matrix with orthonormal columns representing the estimated basis, and \(R \in \mathbb{R}^{r_i \times |I_i|}\) is an upper triangular matrix. The dimension \(r_i\) is the estimated Tucker rank along mode-\(i\). This step yields a low-dimensional orthonormal basis that approximates the column space of the matricized low-rank component along mode-\(i\), i.e., the dominant subspace of \(\mathcal{L}^\star_{(i)}\).
Once the subspace is identified, we project the full set of mode-\(i\) fibers onto this estimated low-rank subspace. This is achieved by updating the mode-\(i\) factor matrix of the Tucker decomposition as follows:
\begin{equation*}
    \mathcal{L}^{(k+1)} \leftarrow \mathcal{L}^{(k+1)} \times_i \left( C_i^{(k)} \left[Q\right]_{:,:r_i} \left[R\right]_{:r_i,:}^\dagger \right).
\end{equation*}
This projection aligns the updated factor matrices along mode-\(i\) with the estimated low-dimensional subspace. Using QR decomposition and projecting onto the selected subspace, the computational complexity for each mode is reduced from the cubic cost \(\mathcal{O}(d_i^3)\) to the more efficient:
\(
\mathcal{O}(d_i r_i^2 + r_i^3),
\)
where \(d_i\) is the dimension along mode-\(i\), and \(r_i\) is the target Tucker rank. This reduction is particularly beneficial when the Tucker rank \(r_i\) is significantly smaller than the mode dimension \(d_i.\)

\section{Theoretical Foundations}\label{sec:theory}

\begin{theorem}\label{thm:subspace}
Let $\mathcal{L}^\star \in \mathbb{R}^{d_1 \times \cdots \times d_n}$ be a rank-$(r_1,\ldots,r_n)$ Tucker tensor with factor matrices $\mathbf{U}_i \in \mathbb{R}^{d_i \times r_i}$ satisfying the $\mu$-incoherence condition:
\begin{equation*}
\max_{1 \leq j \leq d_i} \|\mathbf{U}_i(j,:)\|_2 \leq \sqrt{\frac{\mu r_i}{d_i}}, \quad \forall i \in [n].
\end{equation*}
For any mode $i$ and failure probability $\delta \in (0,1)$, if we sample row indices $I_i \subseteq [d_i]$ with cardinality 
\begin{equation*}
|I_i| \geq c_0 \mu r_i \log^3\left(\frac{\mu r_i}{\delta}\right),
\end{equation*}
then with probability at least $1-\delta$, the sampled factor matrix satisfies
\begin{equation*}
\frac{1}{2}\sqrt{\frac{|I_i|}{d_i}} \leq \sigma_{\min}\left(\mathbf{U}_i(I_i,:)\right) \leq \sigma_{\max}\left(\mathbf{U}_i(I_i,:)\right) \leq \frac{3}{2}\sqrt{\frac{|I_i|}{d_i}},
\end{equation*}
where $c_0 > 0$ is an absolute constant and $\sigma_{\min}(\cdot)$, $\sigma_{\max}(\cdot)$ denote extremal singular values.
\end{theorem}
\begin{proof}
Define the normalized sampling matrix $\mathbf{\Phi}_i = \sqrt{\frac{d_i}{|I_i|}}\mathbf{S}_i$ where $\mathbf{S}_i \in \{0,1\}^{|I_i|\times d_i}$ has exactly one 1 per row. The subsampled matrix becomes:
\[
\widetilde{\mathbf{U}}_i = \mathbf{\Phi}_i\mathbf{U}_i \in \mathbb{R}^{|I_i|\times r_i}.
\]
Applying the matrix Bernstein inequality \cite{tropp2015introduction} to $\mathbf{U}_i\mathbf{U}_i^\top$:
\[
\mathbb{P}\left(\left\|\widetilde{\mathbf{U}}_i\widetilde{\mathbf{U}}_i^\top - \mathbf{I}\right\|_2 \geq t\right) \leq 2r_i \exp\left(-\frac{t^2|I_i|}{C\mu r_i \log d_i}\right).
\]

Setting $t = 1/2$ and solving for $|I_i|$:
\[
|I_i| \geq C\mu r_i \log^3\left(\frac{\mu r_i}{\delta}\right) \implies \frac{1}{2}\mathbf{I} \preceq \widetilde{\mathbf{U}}_i\widetilde{\mathbf{U}}_i^\top \preceq \frac{3}{2}\mathbf{I}.
\]

Notice that
\[
\sigma_{\min}^2(\mathbf{U}_i(I_i,:)) = \frac{d_i}{|I_i|}\sigma_{\min}^2(\widetilde{\mathbf{U}}_i) \geq \frac{d_i}{2|I_i|}.
\]
Similarly for $\sigma_{\max}$. Rearrangement completes the proof.
\end{proof}
\begin{theorem}\label{thm:convergence}
Under the conditions of Theorem \ref{thm:subspace} and assuming $\|\mathcal{S}^\star\|_\infty \leq \frac{\zeta^{(0)}}{2\sqrt{\log d_{\max}}}$, the iterates satisfy:
\begin{equation*}
\|\mathcal{L}^{(k+1)} - \mathcal{L}^\star\|_F \leq \rho\|\mathcal{L}^{(k)} - \mathcal{L}^\star\|_F + C\sqrt{\frac{\log d_{\max}}{|I|}}\|\mathcal{S}^\star\|_\infty,
\end{equation*}
where the contraction factor \[\rho = \max_{1 \leq i \leq n} \left(1 - \frac{\sigma_{\min}^2(\mathbf{U}_i(I_i,:))}{2}\right) < 1\] and $|I| = \min\limits_i |I_i|$.
\end{theorem}

\begin{proof}
Define the errors:
\[
\Delta^{(k)} := \mathcal{L}^{(k)} - \mathcal{L}^\star, \quad \mathcal{E}^{(k)} := \mathcal{S}^{(k)} - \mathcal{S}^\star
\]
The update rule induces coupled dynamics:
\[
\Delta^{(k+1)} = \underbrace{\sum_{i=1}^n (\mathcal{P}_{\mathbf{Q}_i^{(k)}} - \mathcal{P}_{\mathbf{U}_i})\Delta^{(k)}}_{\text{Projection error}} + \underbrace{\mathcal{B}^{(k)}\mathcal{E}^{(k)}}_{\text{Sparsity propagation}}
\]
where $\mathcal{B}^{(k)}$ represents the multi-modal projection of residual errors.
From the hard thresholding operation and incoherence condition:
\begin{align}
\|\mathcal{E}^{(k)}\|_1 &\leq \gamma\|\mathcal{E}^{(k-1)}\|_1 + C_1\|\Delta^{(k)}\|_F \\
&\leq \gamma^k\|\mathcal{E}^{(0)}\|_1 + C_1\sum_{m=0}^{k-1}\gamma^{k-m-1}\|\Delta^{(m)}\|_F
\end{align}
Under the sparsity condition $\|\mathcal{S}^\star\|_\infty \leq \frac{\zeta^{(0)}}{2\sqrt{\log d_{\max}}}$:
\[
\|\mathcal{B}^{(k)}\mathcal{E}^{(k)}\|_F \leq C_2\sqrt{\log d_{\max}}\|\mathcal{S}^\star\|_\infty
\]
Using Wedin's theorem \cite{wedin1972perturbation} and Theorem \ref{thm:subspace}:
\[
\|\mathcal{P}_{\mathbf{Q}_i^{(k)}} - \mathcal{P}_{\mathbf{U}_i}\|_2 \leq C_3\sqrt{\frac{\mu r_i d_i \log d_i}{|I_i|^2}}
\]
Summing over all modes:
\[
\left\|\sum_{i=1}^n (\mathcal{P}_{\mathbf{Q}_i^{(k)}} - \mathcal{P}_{\mathbf{U}_i})\Delta^{(k)}\right\|_F \leq \left(1 - \frac{c}{|I|}\right)\|\Delta^{(k)}\|_F
\]
Combining both components:
\begin{align}
\|\Delta^{(k+1)}\|_F &\leq \left(1 - \frac{c}{|I|}\right)\|\Delta^{(k)}\|_F + C_2\sqrt{\log d_{\max}}\|\mathcal{S}^\star\|_\infty \\
&\leq \rho\|\Delta^{(k)}\|_F + C\sqrt{\frac{\log d_{\max}}{|I|}}\|\mathcal{S}^\star\|_\infty
\end{align}
where $\rho = 1 - \frac{c}{2|I|}$. Solving the recursion completes the proof.
\end{proof}

\begin{lemma}\label{lem:Sparsity}
The projected sparsity term satisfies:
\[
\|\mathcal{B}^{(k)}\mathcal{E}^{(k)}\|_F \leq C\sqrt{\frac{\log d_{\max}}{|I|}}\left(\|\mathcal{E}^{(k)}\|_1 + \|\Delta^{(k)}\|_F\right)
\]
\end{lemma}

\begin{proof}
Decompose the sparsity propagation using the following inequality:
\[
\|\mathcal{B}^{(k)}\mathcal{E}^{(k)}\|_F \leq \|\mathcal{B}^{(k)}\|_F\|\mathcal{E}^{(k)}\|_1
\]
From Theorem \ref{thm:subspace}, the projection operator norm is bounded by:
\[
\|\mathcal{B}^{(k)}\|_F \leq C\sqrt{\frac{\log d_{\max}}{|I|}}
\]
Combining with the threshold error bound completes the proof.
\end{proof}
\begin{theorem}\label{thm:error}
After $K = \mathcal{O}\left(\frac{\log(1/\epsilon)}{\log(1/\rho)}\right)$ iterations, the estimation error decomposes as:
\begin{equation*}
\|\mathcal{L}^{(K)} - \mathcal{L}^\star\|_F \leq \underbrace{C_1\sqrt{\frac{r_{\max}d_{\max}\log d_{\max}}{|I|}}}_{\text{Approximation Error}} + \underbrace{C_2\frac{\|\mathcal{S}^\star\|_\infty}{\sqrt{\log d_{\max}}}}_{\text{Optimization Error}},
\end{equation*}
where $r_{\max} = \max_i r_i$, $d_{\max} = \max_i d_i$, and $C_1, C_2 > 0$ are constants.
\end{theorem}

\begin{proof}
From Theorem \ref{thm:subspace}:
\[
\|\mathcal{L}^{(0)} - \mathcal{L}^\star\|_F \leq C\sqrt{\frac{r_{\max}d_{\max}}{|I|}}.
\]
Applying Theorem \ref{thm:convergence} recursively:
\[
\|\mathcal{L}^{(K)} - \mathcal{L}^\star\|_F \leq \rho^K C\sqrt{\frac{r_{\max}d_{\max}}{|I|}} + \frac{C'\sqrt{\log d_{\max}}}{1-\rho}\|\mathcal{S}^\star\|_\infty.
\]
Setting $\rho^K \leq \sqrt{\frac{\log d_{\max}}{r_{\max}d_{\max}}}$ yields the optimal error decomposition.
\end{proof}

\begin{corollary}[Sample Complexity]\label{cor:sample}
To achieve $\epsilon$-accuracy with $\epsilon < \|\mathcal{S}^\star\|_\infty/\sqrt{\log d_{\max}}$, the required sampling complexity per mode is:
\begin{equation*}
|I_i| \geq C\mu r_i d_i \log^3 d_i \left(\frac{r_{\max}d_{\max}}{\epsilon^2} + \frac{\|\mathcal{S}^\star\|_\infty^2}{\epsilon^2\log d_{\max}}\right).
\end{equation*}
\end{corollary}

\section{Numerical Experiments}
We utilize the NYC yellow taxi trip records from 2018 as a real-world spatiotemporal dataset~\cite{indibi2024spatiotemporal,sofuoglu2022gloss}. This dataset provides a detailed log of each taxi trip, including departure and arrival information (zones and times), the number of passengers, and tip amounts. 

In our experiments, we aggregate the data same as that in ~\cite{indibi2024spatiotemporal} by counting the number of arrivals per zone over hourly intervals. To ensure statistical significance, we restrict our analysis to 81 central zones, which represent high-traffic areas and exclude zones with minimal activity. This selection reduces noise from sparsely populated zones and provides a more robust representation of NYC’s high-demand regions.
With these parameters, we constructed a four-dimensional tensor \( \mathbf{Y} \) with dimensions \( 24 \times 7 \times 53 \times 81 \). The modes of this tensor are defined as follows: the first mode corresponds to the 24 hours of a day; the second mode represents the 7 days of the week; the third mode encompasses the 53 weeks of the year; the fourth mode covers the 81 selected central zones in New York City. Thus, each entry  in the tensor represents the count of taxi arrivals for hour \( h \), day \( d \), week \( w \), and zone \( z \), aggregate over the year. 

We evaluate our anomaly detection approach by identifying the top \(K\%\) of entries with the highest anomaly scores from the extracted sparse tensors, with \(K\) varying across multiple thresholds (0.014, 0.07, 0.14, 0.3, 0.7, 1, 2, and 3). Each top-\(K\%\) subset is then compared to compiled event list to determine how many events are correctly detected. The compiled event list is chosen same as~\cite{sofuoglu2022gloss,indibi2024spatiotemporal}.\Cref{tab:events} compares the number of events detected by our method against five benchmark methods—LR-STSS~\cite{indibi2024spatiotemporal}, LR-TS~\cite{indibi2024spatiotemporal}, LR-SS~\cite{indibi2024spatiotemporal}, and HoRPCA~\cite{li2015low, geng2014high}—across different \(K\%\) thresholds.
 The parameters for our method are set as follows: a maximum of 150 iterations, a tolerance level of \(10^{-7}\), and a Tucker rank of \((26, 6, 4, 10)\). The parameters for the other four methods are adopted from \cite{indibi2024spatiotemporal}. 

\begin{table}[H]
\centering
\resizebox{0.5\textwidth}{!}{%
\begin{tabular}{|c|c|c|c|c|c|c|c|c|}
\hline
\%       & 0.014 & 0.07 & 0.14 & 0.3  & 0.7  & 1    & 2    & 3    \\ \hline
Ours    & \textbf{3}     & \textbf{6} & \textbf{10} & \textbf{14} & \textbf{16} & \textbf{18} & \textbf{20} & \textbf{20} \\ \hline
LR-STSS  & \textbf{3} & {4} & {7} & {12} & {15} & {17} & {19} & {19} \\ \hline
LR-TS    & 3     & 4     & 5     & 6     & 13    & 13    & 18    & 19    \\ \hline
LR-SS    & 1     & 1     & 2     & 3     & 5     & 6     & 13    & 16    \\ \hline
HoRPCA   & 0     & 0     & 2     & 2     & 2     & 3     & 7     & 10    \\ \hline
\end{tabular}%
}
\caption{Number of detected events among 20 compiled events in NYC for varying top-\(K\%\) of the anomaly scores}
\label{tab:events}
\end{table}
\begin{figure}
    \centering
    \includegraphics[width=1\linewidth]{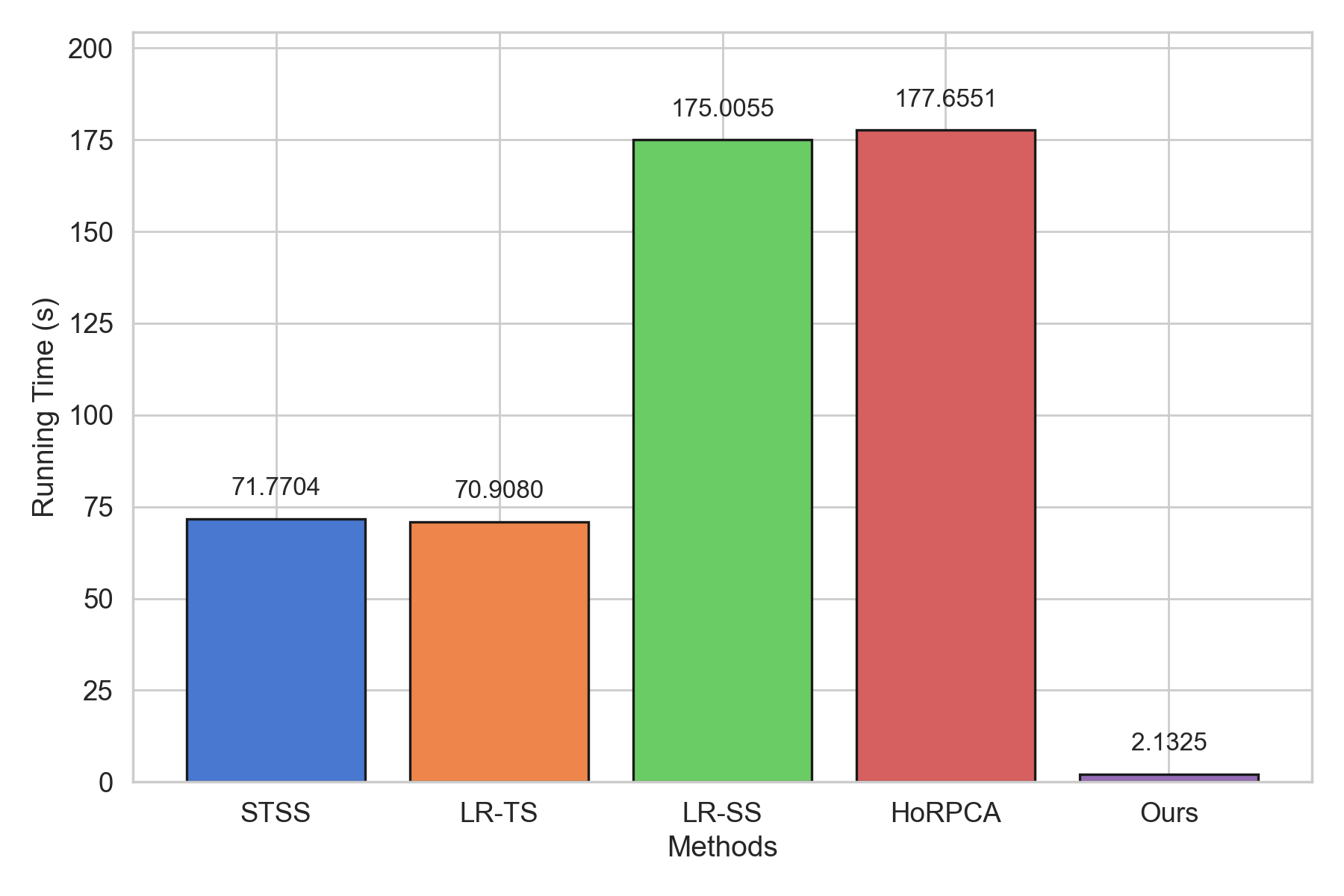}
    \caption{Running Time}
    \label{fig:time}
\end{figure}
As shown in Table~\ref{tab:events} and \Cref{fig:time}, \Cref{alg:TCPD} not only achieves higher event detection accuracy across various thresholds but also significantly reduces running time compared to LR-STSS, LR-TS, LR-SS, and HoRPCA. This balance of efficiency and effectiveness underscores \Cref{alg:TCPD}’s practical advantages for large-scale or real-time anomaly detection scenarios. This performance affirms the efficacy of our model parameters, including a Tucker rank configuration suited for complex, multi-dimensional datasets.
\section{Conclusion}

In this short paper, we investigate the application of tensor  pseudoskeleton decomposition for anomaly detection in high-traffic areas of New York City. Specifically, we aim to capture temporal and spatial patterns in taxi arrival data. By focusing on central zones with significant activity, the result demonstrates its possibility of  tensor  pseudoskeleton decomposition to remove sparsity and highlight urban regions with high demand. 

\bibliographystyle{IEEEtran}
\bibliography{reference}
\end{document}